# Scaling Up Coordinate Descent Algorithms for Large $\ell_1$ Regularization Problems

**Chad Scherrer**  CHAD.SCHERRER@PNNL.GOV
Pacific Northwest National Laboratory

**Mahantesh Halappanavar**  MAHANTESH.HALAPPANAVAR@PNNL.GOV
Pacific Northwest National Laboratory

**Ambuj Tewari**  AMBUJ@CS.UTEXAS.EDU
University of Texas at Austin

**David Haglin**  DAVID.HAGLIN@PNNL.GOV
Pacific Northwest National Laboratory

## Abstract

We present a generic framework for parallel coordinate descent (CD) algorithms that includes, as special cases, the original sequential algorithms Cyclic CD and Stochastic CD, as well as the recent parallel Shotgun algorithm. We introduce two novel parallel algorithms that are also special cases—Thread-Greedy CD and Coloring-Based CD—and give performance measurements for an OpenMP implementation of these.

## 1. Introduction

Consider the $\ell_1$-regularized loss minimization problem

$$\min_{\mathbf{w}} \ \frac{1}{n} \sum_{i=1}^{n} \ell(\mathbf{y}_i, (\mathbf{Xw})_i) + \lambda \|\mathbf{w}\|_1 \ , \qquad (1)$$

where $\mathbf{X} \in \mathbb{R}^{n \times k}$ is the design matrix, $\mathbf{w} \in \mathbb{R}^k$ is a weight vector to be estimated, and the loss function $\ell$ is such that $\ell(y, \cdot)$ is a convex differentiable function for each $y$. This formulation includes Lasso ($\ell(y,t) = \frac{1}{2}(y-t)^2$) and $\ell_1$-regularized logistic regression ($\ell(y,t) = \log(1 + \exp(-yt))$).

In this context, we consider a *coordinate descent* (CD) algorithm to be one in which each iteration performs updates to some number of coordinates of $\mathbf{w}$, and each such update requires traversal of only one column of $\mathbf{X}$.

The goal of the current work is to identify and exploit the available parallelism in this class of algorithms, and to provide an abstract framework that helps to structure the design space.

Our approach focuses on shared-memory architectures, and in particular our experiments use the OpenMP programming model.

This paper makes several contributions to this field. First, we present GenCD, a **Gen**eric **C**oordinate **D**escent framework for expressing parallel coordinate descent algorithms, and we explore how each step should be performed in order to maximize parallelism.

In addition, we introduce two novel special cases of this framework: Thread-Greedy Coordinate Descent and Coloring-Based coordinate descent, and compare their performance to that of a reimplementation of the recent Shotgun algorithm of Bradley et al. (2011), in the context of logistic loss.

A word about notation: We use bold to indicate vectors, and uppercase bold to indicate matrices. We denote the $j$th column of $\mathbf{X}$ by $\mathbf{X}_j$ and the $i$th row of $\mathbf{X}$ by $\mathbf{x}_i^T$. We use $\mathbf{e}^j$ to denote the vector in $\mathbb{R}^k$ consisting of all zeros except for a one in coordinate $j$. We assume the problem at hand consists of $n$ samples and $k$ features.

Table 1. Arrays stored by GenCD

| Name | Dim | Description | Step |
|---|---|---|---|
| $\boldsymbol{\delta}$ | $k$ | proposed increment | Propose |
| $\boldsymbol{\varphi}$ | $k$ | proxy | Propose |
| $\mathbf{w}$ | $k$ | weight estimate | Update |
| $\mathbf{z}$ | $n$ | fitted value | Update |

## 2. GenCD: A Generic Framework for Parallel Coordinate Descent

We now present GenCD, a generic coordinate descent framework. Each iteration computes *proposed increments* to the weights for some subset of the coordinates, and then accepts a subset of these proposals, and modifies the weights accordingly. Algorithm 1 gives a high-level description of GenCD.

---
**Algorithm 1** GenCD

   **while** not converged **do**
      *Select* a set of coordinates $J$
      *Propose* increment $\boldsymbol{\delta}_j, j \in J$    // parallel
      *Accept* some subset $J' \subseteq J$ of the proposals
      *Update* weight $\mathbf{w}_j$ for all $j \in J'$    // parallel

---

In the description of GenCD, we refer to a number of different arrays. These are summarized in Table 1. For some algorithms, there is no need to maintain a physical array in memory for each of these. For example, it might be enough for each thread to have a representation of the proposed increment $\boldsymbol{\delta}_j$ for the column $j$ it is currently considering. In such cases the array-oriented presentation serves only to benefit the uniformity of the exposition.

### 2.1. Step One: Select

We begin by selecting $J$ coordinates for consideration during this major step in an iteration of GenCD. The selection criteria differs for variations of CD techniques.

There are some special cases at the extremes of this selection step. Sequential algorithms like cyclic CD (CCD) and stochastic CD (SCD) correspond to selection of a singleton, while parallel "full greedy" CD corresponds to $J = \{1, \cdots, k\}$.

Between these extremes, Shotgun (Bradley et al., 2011) selects a random subset of a given size. Other obvious choices include selection of a random block of coordinates from some predetermined set of blocks.

Though proposals can be computed in parallel, the number of features is typically far greater than the number of available threads. Thus in general, we would like to have a mechanism for scaling the number of proposals according to the degree of available parallelism. As the first step of each iteration, we therefore select a subset $J$ of the coordinates for which proposals will be computed.

### 2.2. Step Two: Propose

Given the set $J$ of selected coordinates, the second step computes a *proposed increment* $\boldsymbol{\delta}_j$ for each $j \in J$. Note that this step does not actually change the weights; $\boldsymbol{\delta}_j$ is simply the increment to $\mathbf{w}_j$ *if it were to be updated*. This is critical, as it allows evaluation over all selected coordinates to be performed in parallel without concern for conflicts.

Some algorithms such as Greedy CD require a way of choosing a strict subset of the proposals to accept. Such cases require computation of a value on which to base this decision. Though it would be natural to compute the objective function for each proposal, we present the more general case where we have a *proxy* for the objective, on which we can base this decision. This allows for cases where the objective function is relatively expensive but can be quickly approximated.

We therefore maintain a vector $\boldsymbol{\varphi} \in \mathbb{R}^k$, where $\boldsymbol{\varphi}_j$ is a proxy for the objective function evaluated at $\mathbf{w} + \boldsymbol{\delta}_j \mathbf{e}^j$, and update $\boldsymbol{\varphi}_j$ whenever a new proposal is calculated for $j$.

Algorithm 2 shows a generic Propose step.

---
**Algorithm 2** Propose step

   **for each** $j \in J$ **do**    // parallel
      Update proposed increment $\boldsymbol{\delta}_j$
      Update proxy $\boldsymbol{\varphi}_j$

---

### 2.3. Step Three: Accept

Ideally, we would like to accept as many proposals as possible in order to avoid throwing away work done in the proposal step. Unfortunately, as Bradley et al. (2011) show, correlations among features can lead to divergence if too many coordinates are updated at once. So in general, we must restrict to some subset $J' \subseteq J$. We have not yet explored conditions of convergence for our new Thread-Greedy algorithm (see Section 7) but find robust convergence experimentally.

In some algorithms (CCD, SCD, and Shotgun), features are selected in a way that allows all proposals to be accepted. In this case, an implementation can simply bypass the Accept step.

## 2.4. Step Four: Update

After determining which proposals have been accepted, we must update according to the set $J'$ of proposed increments that have been accepted.

In some cases, the proposals might have been computed using an approximation, in which case we might like to perform a more precise calculation for those that are accepted.

As shown in Algorithm 3, this step updates the weights, the fitted values, and also the derivative of the loss.

---

**Algorithm 3** Update step

**for each** $j \in J'$ **do**  // parallel
    Improve $\boldsymbol{\delta}_j$
    $\mathbf{w}_j \leftarrow \mathbf{w}_j + \boldsymbol{\delta}_j$
    $\mathbf{z} \leftarrow \mathbf{z} + \boldsymbol{\delta}_j \mathbf{X}_j$  // atomic

---

Note that each iteration of the **for** loop can be done in parallel. The $\mathbf{w}_j$ updates depend only upon the improved $\boldsymbol{\delta}_j$. While the updates to $\mathbf{z}$ have the potential for collisions if $\mathbf{X}_{ij_1} = \mathbf{X}_{ij_2}$ for some distinct $j_1, j_2 \in J'$, this is easily avoided by using atomic memory updates available in OpenMP and other shared memory platforms.

## 3. Approximate Minimization

As described above, the Propose step calculates a proposed increment $\boldsymbol{\delta}_j$ for each $j \in J$. For a given $j$, we would ideally like to calculate

$$\hat{\delta} = \underset{\delta}{\operatorname{argmin}} \, F(\mathbf{w} + \delta \mathbf{e}^j) + \lambda |\mathbf{w}_j + \delta| , \quad (2)$$

where $F$ is the smooth part of the objective function,

$$F(\mathbf{w}) = \frac{1}{n} \sum_{i=1}^{n} \ell(\mathbf{y}_i, (\mathbf{X}\mathbf{w})_i) . \quad (3)$$

Unfortunately, for a general loss function, there is no closed-form solution for full minimization along a given coordinate. Therefore, one often resorts to one-dimensional numerical optimization.

Alternatively, approximate minimization along a given coordinate can avoid an expensive line search. We present one such approach that follows Shalev-Shwartz & Tewari (2011).

First, note that the gradient and Hessian of $F$ are

$$\nabla F(\mathbf{w}) = \frac{1}{n} \sum_i \ell'(\mathbf{y}_i, (\mathbf{X}\mathbf{w})_i) \mathbf{x}_i$$

$$\mathbf{H}(\mathbf{w}) = \frac{1}{n} \sum_i \ell''(\mathbf{y}_i, (\mathbf{X}\mathbf{w})_i) \mathbf{x}_i \mathbf{x}_i^T .$$

Here $\ell'$ and $\ell''$ denote differentiation with respect to the first variable.

### 3.1. Minimization for squared loss

For the special case of squared loss, the exact minimizer along a coordinate can be computed in closed from. In this case, $\ell''(y, t) \equiv 1$, the Hessian $\mathbf{H}$ is constant, and we have the second-order expansion

$$F(\mathbf{w} + \boldsymbol{\delta}) = F(\mathbf{w}) + \langle \nabla F(\mathbf{w}), \boldsymbol{\delta} \rangle + \frac{1}{2} \boldsymbol{\delta}^\top \mathbf{H} \boldsymbol{\delta} .$$

Along coordinate $j$, this reduces to

$$F(\mathbf{w} + \delta \mathbf{e}^j) = F(\mathbf{w}) + \nabla_j F(\mathbf{w}) \delta + \frac{\mathbf{H}_{jj}}{2} \delta^2 .$$

As is well known (see, for example, (Yuan & Lin, 2010)), the minimizer of (2) is then given by

$$\hat{\delta} = -\psi \left( \mathbf{w}_j; \frac{\nabla_j F(\mathbf{w}) - \lambda}{\mathbf{H}_{jj}}, \frac{\nabla_j F(\mathbf{w}) + \lambda}{\mathbf{H}_{jj}} \right) , \quad (4)$$

where $\psi$ is the clipping function

$$\psi(x; a, b) = \begin{cases} a & \text{if } x < a \\ b & \text{if } x > b \\ x & \text{otherwise} . \end{cases}$$

Note that this is equivalent to

$$\hat{\delta} = s_{\lambda/\mathbf{H}_{jj}} \left( \mathbf{w}_j - \frac{\nabla_j F(\mathbf{w})}{\mathbf{H}_{jj}} \right) - \mathbf{w}_j ,$$

where $s$ is the "soft threshold" function

$$s_\tau(x) = \operatorname{sign}(x) \cdot (|x| - \tau)_+ ,$$

as described by Shalev-Shwartz & Tewari (2011).

### 3.2. Bounded-convexity loss

More generally, suppose there is some $\beta$ such that $\frac{\partial^2}{\partial z^2} \ell(y, z) \leq \beta$ for all $y, z \in \mathbb{R}$. This condition holds for squared loss ($\beta = 1$) and logistic loss ($\beta = 1/4$). Then, let

$$\tilde{F}_\mathbf{w}(\mathbf{w} + \boldsymbol{\delta}) = F(\mathbf{w}) + \langle \nabla F(\mathbf{w}), \boldsymbol{\delta} \rangle + \frac{\beta}{2} \boldsymbol{\delta}^\top \boldsymbol{\delta} .$$

In particular, along any given coordinate $j$, we have

$$\tilde{F}_{\mathbf{w}}(\mathbf{w} + \delta \mathbf{e}^j) = F(\mathbf{w}) + \nabla_j F(\mathbf{w})\delta + \frac{\beta}{2}\delta^2 \quad (5)$$
$$\geq F(\mathbf{w} + \delta \mathbf{e}^j) \ . \quad (6)$$

By (4), this upper bound is minimized at

$$\tilde{\delta} = -\psi\left(\mathbf{w}_j; \frac{\nabla_j F(\mathbf{w}) - \lambda}{\beta}, \frac{\nabla_j F(\mathbf{w}) + \lambda}{\beta}\right) \ . \quad (7)$$

Because the quadratic approximation is an upper bound for the function, updating $\boldsymbol{\delta}_j \leftarrow \tilde{\delta}$ is guaranteed to never increase the objective function.

### 3.3. A proxy for the objective

In cases where $J'$ is a strict subset of $J$, there must be some basis for the choice of which proposals to accept. The quadratic approximation provides a way of quickly approximating the potential effect of updating a single coordinate. For a given weight vector $\mathbf{w}$ and coordinate $j$, we can approximate the decrease in the objective function by

$$\varphi_{\mathbf{w}}^j = \tilde{F}_{\mathbf{w}}(\mathbf{w} + \tilde{\delta}\mathbf{e}^j) - \tilde{F}_{\mathbf{w}}(\mathbf{w}) \quad (8)$$
$$= \frac{\beta}{2}\tilde{\delta}^2 + \nabla_j F(\mathbf{w})\tilde{\delta} + \lambda(|\mathbf{w}_j + \tilde{\delta}| - |\mathbf{w}_j|) \ . \quad (9)$$

## 4. Experiments

### 4.1. Algorithms

At this point we discuss a number of algorithms that are special cases of the GenCD framework. Our intent is to demonstrate some of the many possible instantiations of this approach. We perform convergence and scalability tests for these cases, but do not favor any of these algorithms over the others.

For current purposes, we focus on the problem of classification, and compare performance across algorithms using logistic loss. All of the algorithms we tested benefited from the addition of a line search to improve the weight increments in the Update step. Our approach to this was very simple: For each accepted proposal increment, we perform an additional 500 steps using the quadratic approximation.

SHOTGUN is a simple approach to parallel CD introduced by Bradley et al. (2011). In terms of GenCD, the Select step of Shotgun chooses a random subset of the columns, and the Accept step accepts every proposal. This makes implementation very simple, because it eliminates the need for computation of a proxy, and allows for fusion of the Propose and Update loops.

**Algorithm 4** Proposal via approximation

  **for each** $j \in J$ **do**                      // parallel
    $g \leftarrow \langle \ell'(\mathbf{y}_i, \mathbf{z}_i), \mathbf{X}_j\rangle /n$         // thread-local
    $\boldsymbol{\delta}_j \leftarrow -\psi\left(\mathbf{w}_j; \frac{g-\lambda}{\beta}, \frac{g+\lambda}{\beta}\right)$
    $\boldsymbol{\varphi}_j \leftarrow \frac{\beta}{2}\boldsymbol{\delta}_j^2 + g\boldsymbol{\delta}_j + \lambda(|\mathbf{w}_j + \boldsymbol{\delta}_j| - |\mathbf{w}_j|)$

Bradley et al. (2011) show that convergence of Shotgun is guaranteed only if the number of coordinates selected is at most $P^* = \frac{k}{2\rho}$, where $k$ is the number of columns of $\mathbf{X}$, and $\rho$ is the spectral radius ($\equiv$ maximal eigenvalue) of $\mathbf{X}^\top\mathbf{X}$. For our experiments, we therefore use $P^*$ columns for this algorithm.

Bradley et al. (2011) suggest beginning with a large regularization parameter, and decreasing gradually through time. Since we do not implement this, our results do not apply to Shotgun variations that use this approach.

In our novel THREAD-GREEDY algorithm, the Select step chooses a random set of coordinates. Proposals are calculated in parallel using the approximation described in Equation (7) of Section 3.2, with each thread generating proposals for some subset of the coordinates. Each thread then accepts the best of the proposals it has generated.

We have not yet determined conditions under which this approach can be proven to converge, but our empirical results are encouraging.

Similarly, GREEDY selects all coordinates, and uses the computed proxy values to find the best for each thread. However, rather than accept one proposal for each thread, proxy values are compared across threads, and only the single best is chosen in the Accept step. Dhillon et al. (2011) provide an analysis of convergence for greedy coordinate descent.

COLORING is a novel algorithm introduced in this paper. The main idea here is to determine sets of structurally independent features, in order to allow safe concurrent updates without requiring any synchronization between parallel processes (threads). Coloring begins with the preprocessing step, where structurally independent features are identified via *partial distance-2 coloring* of a bipartite graph representation of the feature matrix. In the Select step, a random color (or a random feature) is selected. All the features that have been assigned this color are updated concurrently. Based on the colors, a different number of features could get selected at each iteration.

In contrast to Shotgun, conflicting updates are not lost. However, we note that Coloring is suitable only for sparse matrices with sufficient independence in the structure. Details of this algorithm are provided in Appendix **A**.

Table 2. Specific cases of the GenCD framework.

| Algorithm | Select | Accept |
|---|---|---|
| SHOTGUN | Rand subset | All |
| THREAD-GREEDY | All | Greedy/thread |
| GREEDY | All | Greedy |
| COLORING | Rand color | All |

Since COLORING only allows parallel updates for features with disjoint support, updating a single color is equivalent to updating each feature of that color in sequence. Thus convergence propoerties for COLORING should be very similar to those of cyclic/stochastic coordinate descent.

Table 2 provides a summary of the algorithms we consider.

### 4.2. Implementation

We provide a brief overview of software implementations in this section. All the four algorithms are implemented in C language using OpenMP as the parallel programming model Chapman et al. (2007). We use GNU C/C++ compilers with `-O3 -fopenmp` flags. Work to be done in parallel is distributed among a team of threads using OpenMP `parallel for` construct. We use `static` scheduling of threads where each thread gets a contiguous block of iterations of the `for` loop. Given $n$ iterations and $p$ threads, each thread gets a block of $\frac{n}{p}$ iterations.

Threads running in parallel need to synchronize for certain operations. While all the algorithms need to synchronize in the Update step, Algorithm GREEDY also needs to synchronize during the Select step to determine the best update. We use different mechanisms to synchronize in our implementations. Concurrent updates to vector **z** are made using atomic memory operations. We use OpenMP `atomic` directive for updating real numbers and use the x86 hardware operator `__sync_fetch_and_add` for incrementing integers. We use OpenMP `critical` sections to synchronize threads in the Propose step of GREEDY algorithm. In order to preserve all the updates, we synchronize in the Update step of SHOTGUN. Note that this is different from the SHOTGUN algorithm as proposed in Bradley et al. (2011). Since structurally independent features are updated at a given point of time in the COLORING algorithm, there is no need for synchronization in the Update step.

### 4.3. Platform

Our experimental platform is an AMD Opteron (Magny-Cours) based system with 48 cores (4 sockets with 12 core processors) and 256 GB of globally addressable memory. Each 12-core processor is a multi-chip module consisting of two 6-core dies with separate memory controllers. Each processor has three levels of caches: 64 KB of L1 (data), 512 KB of L2, and 12 MB of L3. While L1 and L2 are private to each core, L3 is shared between the six cores of a die. Each socket has 64 GB of memory that is globally addressable by all four sockets. The sockets are interconnected using the AMD HyperTransport-3 technology[1].

We conduct scaling experiments on this system in powers of two (1, 2, 4, …, 32). Convergence results provided in Figure 1 are on 32 processors for each input and algorithm.

### 4.4. Datasets

We performed tests on two sets of data. For each, we normalized columns of the feature matrix in order to be consistent with algorithmic assumptions.

DOROTHEA is the drug discovery data described by Guyon et al. (2004). Here, each example corresponds to a chemical compound, and each feature corresponds to a structural molecular feature.

The feature matrix is binary, indicating presence or absence of a given molecular feature. There are 800 examples and 100,000 features, with 7.3 nonzeros per feature on average. We estimate $P^*$ to be about 23. Our coloring resulted in a mean color size of 16.

As a response, we are given an indicator of whether a given compound binds to thrombin. There are 78 examples for which this response is positive.

For the DOROTHEA data set, we used a regularization parameter of $\lambda = 10^{-4}$.

REUTERS is the RCV1-v2/LYRL2004 Reuters text data described by Lewis et al. (2004). In this data, each example corresponds to a training document, and each feature corresponds to a term. Values in the training data correspond to term frequencies, transformed using a standard tf-idf normalization.

The feature matrix consists of 23865 examples and 47237 features. It has 1.7 million nonzeros, or 37.2 nonzeros per feature on average. We estimate $P^*$ to be about 800. Our coloring resulted in a mean color

---
[1] Further details on Opteron can be found at http://www.amd.com/us/products/embedded/processors/opteron/Pages/opteron-6100-series.aspx.

Table 3. A summary of data sets.

|  | DOROTHEA | REUTERS |
|---|---|---|
| Samples | 800 | 23865 |
| Features | 100000 | 47237 |
| Nonzeros/feature | 7.3 | 37.2 |
| $P^*$ | 23 | 800 |
| Features/color | 16 | 22 |
| Time to color | 0.7 sec | 1.6 sec |
| Our chosen $\lambda$ | $10^{-4}$ | $10^{-5}$ |
| $\min F(\mathbf{w}) + \lambda\|\mathbf{w}\|_1$ | 0.279512 | 0.165044 |
| Best-fit NNZ | 14182 | 1903 |

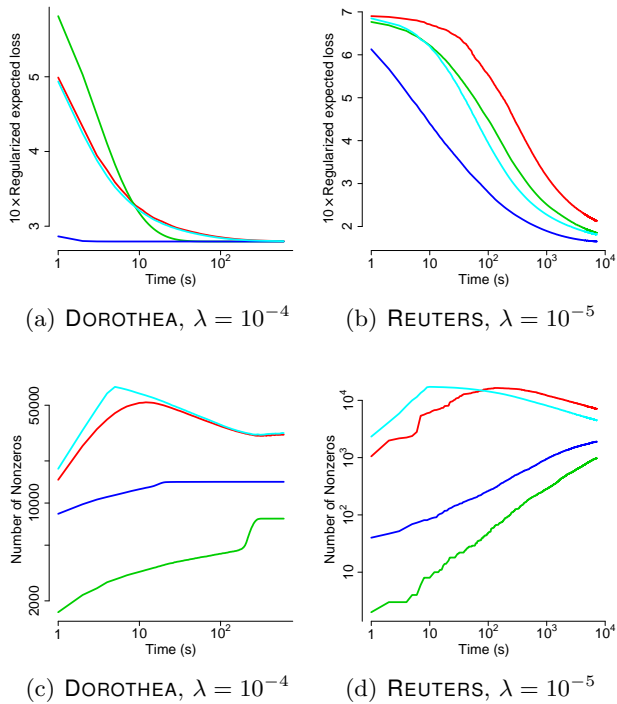

(a) DOROTHEA, $\lambda = 10^{-4}$    (b) REUTERS, $\lambda = 10^{-5}$

(c) DOROTHEA, $\lambda = 10^{-4}$    (d) REUTERS, $\lambda = 10^{-5}$

Figure 1. Convergence results for SHOTGUN, THREAD-GREEDY, GREEDY, and COLORING.

size of 22.

Each document in this data is labeled as belonging to some set of "topics". The most common of these (matching 10786 documents) is CCAT, corresponding to "ALL Corporate-Industrial". Based on this, we used membership in the CCAT topic as a response.

For the REUTERS data set, using a regularization parameter of $\lambda = 10^{-4}$ led to an optimal solution of 0, so we reduced it to $\lambda = 10^{-5}$.

## 5. Results And Discussion

### 5.1. Convergence rates

The results of our convergence experiments are summarized in Figure 1.

First, consider DOROTHEA. Here, Figure 1(a) shows that all four algorithms were very close to convergence by the end of 10 minutes. In particular, for THREAD-GREEDY after the first 224 seconds, both the Objective function and the number of nonzeros (NNZ) were stable.

As expected, GREEDY added nonzeros very slowly, while SHOTGUN and COLORING began by greatly increasing NNZ. Interestingly, though this initially put GREEDY far behind the other algorithms, but by the end of the first ten seconds, it is back on track with SHOTGUN and Coloring.

At around 200 seconds, there is a sudden increase in NNZ for GREEDY. We are uncertain of the source of this phenomenon.

Perhaps most striking is that both SHOTGUN and COLORING tend to begin by greatly increasing the number of nonzeros (NNZ). In the case of a very sparse optimum like that of DOROTHEA with $\lambda = 10^{-4}$, this initial effect is difficult to overcome, and both algorithms are at a disadvantage compared with GREEDY and THREAD-GREEDY. This effect is less dramatic for REUTERS, where the optimal solution has more nonzeros, and we suspect it would disappear entirely for problems with near-dense optima.

Overall, performance of COLORING and SHOTGUN were remarkably similar for both data sets.

### 5.2. Scalability

Figure 2 shows scalability across algorthims, in terms of the number of updates per second. GREEDY makes updates relatively slowly, due to the large amount of work preceding an update in a given iteration. While the proposals can be computed in parallel, the threads must synchronize in order to identify the single best coordinate to update. This synchronization, and the subsequent update in serial, reduces parallel efficiency.

The THREAD-GREEDY has no such synchronization, and allows the number of updates to increase with the number of threads. The only concern we foresee is potential divergence if the number of threads is too large.

On DOROTHEA, COLORING and SHOTGUN have similar scalability. REUTERS has a similar number of feautures per color, but $P^*$ is much higher (800 vs 23 for DOROTHEA). This leads to greater scalability for SHOTGUN in this case, but not for COLORING.

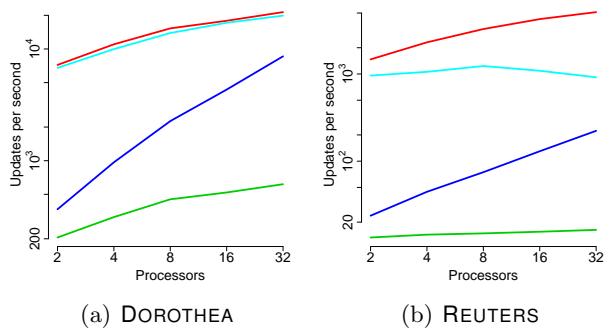

(a) DOROTHEA

(b) REUTERS

Figure 2. Scalability for SHOTGUN, THREAD-GREEDY, GREEDY, and COLORING.

## 6. Related work

The importance of coordinate descent methods for $\ell_1$ regularized problems was highlighted by Friedman et al. (2007) and Wu & Lange (2008). Their insights led to a renewed interest in coordinate descent methods for such problems. Convergence analysis of and numerical experiments with randomized coordinate or block coordinate descent methods can be found in the work of Shalev-Shwartz & Tewari (2011), Nesterov (2010) and Richtárik & Takáč (2011a). Greedy coordinate descent is related to boosting, sparse approximation, and computational geometry. Clarkson (2010) presents a fascinating overview of these connections. Li & Olsher (2009) and Dhillon et al. (2011) apply greedy coordinate descent to $\ell_1$ regularized problems. Recently, a number of authors have looked at parallel implementations of coordinate descent methods. Bradley et al. (2011) consider the approach of ignoring dependencies, and updating a randomly-chosen subset of coordinates at each iteration. They show that convergence is guaranteed up to a number of coordinates that depends on the spectral radius of $\mathbf{X}^\top \mathbf{X}$. Richtárik & Takáč (2011b) present an approach using GPU-accelerated parallel version of greedy and randomized coordinate descent. Yuan & Lin (2010) present an empirical comparison of several methods, including coordinate descent, for solving large scale $\ell_1$ regularized problems. However, the focus is on serial algorithms only. A corresponding empirical study for parallel methods does not exist yet partly because the landscape is still actively being explored. Finally, we note that the idea of using graph coloring to ensure consistency of parallel updates has appeared before (see, for example, (Bertsekas & Tsitsiklis, 1997; Low et al., 2010; Gonzalez et al., 2011)).

## 7. Conclusion

We presented GenCD, a generic framework for expressing parallel coordinate descent algorithms, and described four variations of algorithms that follow this framework. This framework provides a unified approach to parallel CD, and gives a convenient mechanism for domain researchers to tailor parallel implementations to the application or data at hand. Still, there are clearly many questions that need to be addressed in future work.

Our THREAD-GREEDY approach can easily be extended to one that accepts the best $|J'|$ proposals, independently of which thread proposed to update a given coordinate. This would have additional synchronization overhead, and it is an open question whether this overhead could be overcome by the improved Accept step.

Another open question is conditions for convergence of the THREAD-GREEDY algorithm. A priori, one might suspect is would have the same convergence criteria as SHOTGUN. But for DOROTHEA, $P^* = 23$, yet updating a coordinate for each of 32 threads results in excellent convergence.

It is natural to consider extending SHOTGUN by partitioning the columns of the feature matrix into blocks, and then computing a $P_b^*$ for each block $b$. Intuitively, this can be considered a kind of "soft" coloring. It remains to be seen what further connections between these approaches will bring.

Our COLORING algorithm relies on a preprocessing step to color the bipartite graph corresponding to the adjacency matrix. Like most coloring heuristics, ours attempts to minimize the total number of colors. But in the current context, fewer colors does not necessarily correspond with greater parallelism. Better would be to have a more *balanced* color distribution, even if this would require a greater number of colors.